%% file: sample_FG2020.tex
\documentclass[a4paper, 10pt, conference]{ieeeconf}      

\usepackage{FG2020}

\FGfinalcopy 

\IEEEoverridecommandlockouts                              
\usepackage{amsmath} 
\usepackage{graphicx}
\usepackage{makecell}
\def\rot{\rotatebox}
\def\FGPaperID{****} 

\title{\LARGE \bf
Multi-Label Class Balancing Algorithm for Action Unit Detection
}

\author{\parbox{16cm}{\centering
		{\large Jaspar Pahl$^*,^1$, Ines Rieger\thanks{$^*$These authors contributed equally to this paper.}$^*,^1$, Dominik Seuss$^1$}\\
		{\normalsize
			$^1$ Fraunhofer Institute for Integrated Circuits IIS, Erlangen, Germany\\
		\{jaspar.pahl, ines.rieger, dominik.seuss\}@iis.fraunhofer.de}}
	\thanks{This project is funded by the Federal Ministry of Education and Research, grant no. 01IS18056A (TraMeExCo), and the German Research Foundation (PainFaceAnalyzer).}
}
\begin{document}

\ifFGfinal
\thispagestyle{empty}
\pagestyle{empty}
\else
\author{Anonymous FG2020 submission\\ Paper ID \FGPaperID \\}
\pagestyle{plain}
\fi
\maketitle

\begin{abstract}
Isolated facial movements, so-called Action Units, can describe combined emotions or physical states such as pain. As datasets are limited and mostly imbalanced, we present an approach incorporating a multi-label class balancing algorithm. This submission is subject to the Action Unit detection task of the  Affective Behavior Analysis in-the-wild (ABAW) challenge at the IEEE Conference on Face and Gesture  Recognition.

\end{abstract}

\input{introduction.tex}
\input{datasets.tex}

\input{methods.tex}
\input{results.tex}
\input{conclusion.tex}

\bibliographystyle{plain}
\bibliography{bib}
%
%
%
%

\end{document}

%% file: introduction.tex
\section{INTRODUCTION}

Distinct facial movements can be categorized as Action Units (AUs), which are described in the  Facial Action Coding System (FACS) \cite{ekman2002facial}. Based on these AUs, facial expressions such as physical pain \cite{kunz_facial_2019} or the basic emotions anger, disgust, fear, happiness, sadness, and surprise \cite{ekman2003darwin} can be described.

There are several approaches to automatically detect the occurrence and intensity of AUs \cite{martinez2017automatic, rieger2020Verifying}. Deep learning models are one of the most successful approaches, but require large amounts of data to perform well in training and testing. The manual annotation time for AU occurrences and intensity is high and must be conducted by expert FACS-coders. Thus, the training data is limited. Furthermore, AU datasets are mostly imbalanced as different AUs do not occur with the same frequency for different facial expressions. 

We propose a training pipeline with an optimized augmentation approach to compensate for imbalanced AU occurrences  and the data limitations.



%% file: datasets.tex
\section{Datasets\label{datasets}}

We use the Aff-Wild2 \cite{kollias2018aff, kollias2019expression,kollias2018multi, kollias2019deep} as training, validation and test dataset and as additional training datasets the Emotionet \cite{emotionet}, the Actor Study \cite{seuss2019emotion} and the Extended Cohn-Kanade (CK+) \cite{kanade2000comprehensive,lucey2010extended}.

The Aff-Wild2 database is the largest in-the-wild database consisting of videos and is partly annotated for the three tasks valence-arousal estimation \cite{zafeiriou2017aff}, basic expression recognition \cite{kollias2017recognition} and Action Unit detection. All videos were scraped from YouTube. For Action Unit detection, 56 videos with  63  subjects  (32  males and  31  females) were annotated by experts. In total, 398,835 frames are annotated. This dataset is for non-commercial research purposes only.

The Emotionet database consists of approximately one million images scraped from the web using emotive keywords. Benitez-Quiroz et al. \cite{fabian2016emotionet} analyzed the occurrence and intensity of AUs in 90\% of the images automatically. Experts FACS-coded 10\% of the dataset manually. We only use the manual encoded part of the dataset for our experiments. This dataset is for non-commercial research purposes only.

The Actor Study contains sequences of 21 actors, in total 68 minutes of video filmed from different views and camera speed. Each actor had the task to display specific Action Units in different intensities and had to respond to scenarios and enactments. The frames are annotated with Action Units and their corresponding intensities. We use the center view of the low speed camera for our approach. This dataset was recently published and will be made publicly available for commercial research.

The CK+ dataset contains 593 videos of 123 subjects. The sequences show the facial expressions from neutral to strong. All expressions are posed. The presence or absence of the Action Units is coded for the peak frames, the strongest expression, only. This dataset is for non-commercial research purposes only.\\

%% file: methods.tex
\section{Methods\label{methods}}



\subsection{Pre-processing\label{Preprocessing}}

For both the AffWild training and validation set we use the provided cropped and aligned images of pixel size 112 x 112 in order to eliminate error sources and to make our approach easier to compare to those of other teams. The remaining training sets had no pre-cropped images available. For the CK+ dataset and the Actor Study dataset the Sophisticated High-speed Object Recognition Engine SHORE\texttrademark~\cite{kublbeck2006face} was used to crop images of the same size. The Emotionet dataset was cropped using OpenFace~\cite{amos2016openface}. All images were converted to gray scale and pixel values were normalized in a range of $[0,1]$.
Combining all datasets from Section~\ref{datasets} yields a highly imbalanced dataset, which can be seen under the name of non-aug in Table~\ref{tab_aus}.
Imbalanced datasets cause problems in the detection of under-represented Action Units. Therefore, we implemented a multi-label class balancing algorithm which increases the allowed number of augmentations in the under-represented Action Units. 
In a multi-class problem, the occurrences of a class cannot be increased by simply augmenting a single image multiple times because the potential other classes in the image would be increased by the same amount. Therefore, we formulate the class balancing as an optimization problem:

\begin{subequations}
	\begin{equation}
	f(n_{u}) = \sum \mid z - \sum^{N}_{i} \frac{z_{i}}{N}  \mid + \lambda \ Var \left( \frac{n_{u}}{n_{0}}\right)
	\end{equation}
	\begin{equation}
	z = \sum_{r}  n_{u} * y_{u}
	\end{equation}
\end{subequations}

where $n_{u}$ is the vector of number of occurrences of the images with a unique Action Unit combination and $y_{u}$ is the respective one-hot-encoded array of labels for the unique groups. The operator $*$ between $n_{u}$ and $y_{u}$ denotes a scalar multiplication for each row $r$. The variable $n_{0}$ is the vector with the original occurrences in the dataset non-aug, and $\lambda$ is a weighting parameter. Furthermore, the search domain for $n_{u}$ has been restricted as
\begin{equation}
n_{0} < n_{u} < 10 \ n_{0}
\end{equation}
As such, we can control the number of additional augmentations while stopping the optimizer from increasing the relative growth of a single unique label combination too much. The latter would cause problems due to too many augmentations on a single image, which potentially leads to overfitting. \\
The augmentations were implemented using the imgaug library~\cite{imgaug}. We used Flipping, Gaussian Blur, Linear Contrast, Additive Gaussian Noise, Multiply, and Perspective Transform.\\
In Table~\ref{tab_aus} the name aug-1 refers to the set with augmentations as described by the upper algorithm.
%
\begin{table}[b]
	\centering

	\caption{Label distribution of the original training datasets and the augmented dataset	\label{tab_aus}}
\begin{tabular}{r||r|r|r|r||r|r}
		\rot{90}{Dataset} & \rot{90}{AffWild} & \rot{90}{\makecell{10\%\\Emotionet}} &\rot{90}{CK+}  & \rot{90}{Actorstudy} & \rot{90}{\makecell{total train sets \\ non-aug}} & \rot{90}{\makecell{total augmented \\ train sets aug-1}} \\
		\hline 
		\hline 
		\rule{0pt}{10pt}
		AU01 & 47548 & 589 & 2964 & 11079 & 62180 & 62874 \\ 
		\hline 
		\rule{0pt}{10pt}
		AU02 & 2271 & 0 & 1875 & 8553 & 12699 & 24754 \\
		\hline 
		\rule{0pt}{10pt}
		AU04 & 32387 & 676 & 3753 & 11393 & 48209 & 48365 \\ 
		\hline 
		\rule{0pt}{10pt}
		AU06 & 9290 & 0 & 2174 & 6991 & 18455 & 33091 \\ 
		\hline 
		\rule{0pt}{10pt}
		AU12 & 22964 & 0 & 2475 & 7798 & 33237 & 35417 \\
		\hline
		\rule{0pt}{10pt} 
		AU15 & 1537 & 0 & 1654 & 2115 & 5306 & 17748 \\
		\hline 
		\rule{0pt}{10pt}
		AU20 & 3490 & 0 & 1406 & 4704 & 9600 & 25681 \\
		\hline 
		\rule{0pt}{10pt}
		AU25 & 7463 & 1780 & 5359 & 13573 & 28175 & 35416 \\ 
		\hline 
		\hline
		\rule{0pt}{10pt}
		\makecell{total \\ images}  & 232842 & 20160 & 10724 & 58825 & 322551 & 378041 \\ 
	\end{tabular} 
\end{table}


\subsection{Training}
Our backbone is a parameter-reduced 18-layer ResNet (ResNet18) \cite{rieger2019towards} based on He et al. \cite{he2016deep}. Small changes were made in the output layer which uses a sigmoid function and in the activation functions, which are ReLU units. The threshold for a positive detection is 0.5. We use a f1 loss function with an Adam-optimizer and a learning rate of 0.0001. These parameters have been determined by a grid search.

%% file: results.tex
\section{Results}







Table \ref{tab_results_aus} shows the results on the Aff-Wild2 validation dataset. Our results show that augmentation can counteract an imbalanced dataset. Our model trained on the augmented dataset aug-1 performs better at detecting all Action Units equally. Our trained models supersede the baseline model~\cite{kollias2020analysing}.

\begin{table}[h]
	\centering
	\caption{Results on the Aff-Wild2 validation dataset. The results of the AUs are measured in F1 Score. Best results are in bold.	\label{tab_results_aus}}
	\begin{tabular}{c||c|c|c}
		Model & baseline\cite{kollias2020analysing} & ResNet18 (ours) & ResNet18 (ours) \\ 
		Train Set & AffWild2 & non-aug & aug-1 \\ 
		\hline 
		\hline 
		\rule{0pt}{10pt}
		AU01 & - & 0.64 & \textbf{0.71}\\
		\hline
		\rule{0pt}{10pt}
		AU02 & - & 0.00 & 0.00\\
		\hline
		\rule{0pt}{10pt}
		AU04 & - &\textbf{0.47} & 0.36\\
		\hline
		\rule{0pt}{10pt}
		AU06 & - & 0.21 & \textbf{0.33}\\
		\hline
		\rule{0pt}{10pt}
		AU12 & - & 0.39 & \textbf{0.41}\\
		\hline
		\rule{0pt}{10pt}
		AU15 & - &0.00 & \textbf{0.10}\\
		\hline
		\rule{0pt}{10pt}
		AU20 & - & 0.00 & \textbf{0.04}\\
		\hline
		\rule{0pt}{10pt}
		AU25 & - &0.00 & 0.00\\
		\hline
		\rule{0pt}{10pt}
		F1Macro & - & 0.21 & \textbf{0.24}\\
		\hline
		\rule{0pt}{10pt}
		$\frac{F1+Acc}{2}$ & 0.31& 0.58 & \textbf{0.59}\\
	\end{tabular} 
\end{table}

%% file: conclusion.tex
\section{Conclusion}
In this paper we present an approach based on a multi-label class balancing algorithm as a pre-processing step to overcome the imbalanced occurrences of Action Units in the training dataset. We trained a ResNet with 18 layers and could show an improvement of the augmented training dataset in contrast to the original dataset.